\documentclass[11pt,a4paper]{article}
\usepackage[hyperref]{emnlp2020}
\usepackage{times}
\usepackage{latexsym}

\usepackage{microtype}
\usepackage{multirow}
\usepackage{csvsimple}
\usepackage{hyperref}
\usepackage{graphicx}
\usepackage{amssymb}
\usepackage{xcolor}
\usepackage{multicol}
\usepackage{algorithm}
\usepackage[noend]{algpseudocode}

\newcounter{notecounter}
\newcommand{\enotesoff}{\long\gdef\enote##1##2{}}
\newcommand{\enoteson}{\long\gdef\enote##1##2{{
\stepcounter{notecounter}
{\large\bf
\hspace{1cm}\arabic{notecounter} $<<<$ ##1: ##2
$>>>$\hspace{1cm}}}}}
\enoteson
\enotesoff

\aclfinalcopy
\algrenewcommand\algorithmicindent{0.5em}

\def\figref#1{Figure~\ref{fig:#1}}
\def\figlabel#1{\label{fig:#1}\label{p:#1}}

\def\tabref#1{Table~\ref{tab:#1}}
\def\tablabel#1{\label{tab:#1}\label{p:#1}}

\def\secref#1{\S\ref{sec:#1}}
\def\seclabel#1{\label{sec:#1}}
\def\eqref#1{Eq.~\ref{eqn:#1}}

\long\def\eat#1{\ignorespaces}

\newif\ifcomment\commenttrue

\ifcomment
  \newcommand{\prettycomment}[3]{\colorbox{#1}{\parbox{.8\linewidth}{#2: #3}}}
\else
  \newcommand{\prettycomment}[3]{}
\fi

\makeatletter
\newcommand{\printfnsymbol}[1]{%
  \textsuperscript{\@fnsymbol{#1}}%
}
\makeatother

\title{Are Pretrained Language Models Symbolic Reasoners Over Knowledge?}
\author{Nora Kassner\thanks{*equal contribution} ,  Benno Krojer\printfnsymbol{1}, Hinrich Sch\"utze \\
  Center for Information and Language Processing (CIS) \\
  LMU Munich, Germany \\
  \texttt{kassner@cis.lmu.de}}

\date{}

\begin{document}

\maketitle

\begin{abstract}
How can pretrained language models (PLMs) learn factual
knowledge from the training set? We investigate the two 
most important mechanisms: reasoning and memorization.  Prior
work has attempted to quantify the number of facts PLMs
learn, but we present, using synthetic data, the first study
that investigates the causal relation between facts present in
training and facts learned by the PLM. 
For reasoning, we
show that PLMs seem to learn to apply some symbolic reasoning
rules correctly but  struggle with others, including two-hop reasoning.
Further analysis suggests that even the application of
learned reasoning
rules is flawed.
For memorization, we identify schema conformity (facts
systematically supported by other facts) and frequency as
key factors for its success.

\end{abstract}

\enote{hs}{

  issues

  \begin{itemize}
    \item $\times$ lines: how many items?
  \item negation
  \item instance: need to clarify?
    \item sym inv  neg: anzahl additional facts?
  \end{itemize}

  }

\enote{nk}{To do: 1) make pseudo-code pretty 2) put in values 3) cite logicNN 4) maybe add learning curves}
\enote{nk}{Is is clear how our multi-hop setting is more difficult than prior work!!}

\section{Introduction}

 \begin{table*}[!htbp]
  \small
  \centering
\begin{tabular}{ll|l|l}
Rule&& Definition& Example\\ \hline
EQUI & Equivalence &($e, r, a$) $\iff$ ($e, s, a$)  & (bird, can, fly) $\iff$ (bird, is able to, fly) \\
  SYM &Symmetry &  $(e, r, f) \iff (f,r, e)$  &
(barack, married, michelle) $\iff$ (michelle, married, barack)\\
INV &Inversion & ($e, r, f$) $\iff$ ($f, s, e$) & (john, loves, soccer)
$\iff$ (soccer, thrills, john) \\

NEG & Negation &($e, r, a$) $\iff$ ($e$, \mbox{not} $r$, $b$)  & (jupiter, is, big)  $\iff$ (jupiter, is not, small) \\
IMP & Implication &($e, r, a$) $\Rightarrow$  ($e, s, b$), ($e, s, c$),...  & (dog, is, mammal)  $\Rightarrow$ (dog, has, hair), (dog, has, neocortex), ...\\
COMP & Composition & ($e, r, f$) $\wedge$ ($f, s, g$) $\Rightarrow$ ($e, t, g$) & (tiger, faster than, sheep)   $\wedge$ (sheep, faster than, snail)   \\
&&&$\Rightarrow$ (leopard, faster than, snail)   with $r=s=t$ 
\end{tabular}
\caption{\tablabel{patterns}The six symbolic rules we
  investigate (cf.\ \cite{Nayyeri2019LogicENNAN})  with an example in natural language for entities $e,f,g \in {E}$, relations $r,s,t \in {R}$ and attributes $a,b,c \in {A}$.}
\end{table*}

Pretrained language models (PLMs) like BERT
\cite{devlin-etal-2019-bert}, GPT-2 \cite{radford2019language} and RoBERTa \cite{DBLP:journals/corr/abs-1907-11692} have emerged as universal tools
that capture a diverse range of linguistic and
-- as more and more evidence suggests -- factual knowledge  \cite{petroni2019language, radford2019language}.
\enote{nk}{Much of recent work employs PLMs in a blackbox setting on downstream tasks. Some analysis focuses on 
Specially relevant for downstream tasks leveraging knowledge indirectly. add PLMs blackbox use for downstream task. Inner workings not understood. also relevant for downstream tasks leveraging knowledge indirectly}

Recent work on knowledge captured by PLMs is focused on
probing, a methodology that identifies the set of facts a
PLM has command of. But little is understood about how this
knowledge is acquired during pretraining and why.
\enote{hs}{is this the best place of this sentence? it's an
  important consideration. But the preceding sentence gives
  gthe main motivation of the paper, the followin sentence
  then says what we do about it. I don't see waht this
  sentence contributes to this arguemnt
During the
course of pretraining, PLMs see more data than any human
could read in a lifetime, an amount of knowledge that
surpasses the storage capacity of today's models.} We analyze
the ability of PLMs to acquire factual knowledge focusing on
two mechanisms: reasoning and memorization.  We pose the
following two questions:
\textbf{a) Symbolic reasoning:} Are PLMs able to infer knowledge not seen explicitly during pretraining?
\textbf{b) Memorization:}
Which factors result in successful memorization of a fact by PLMs?

We conduct our study
by pretraining BERT from scratch on
synthetic corpora. The corpora are composed of short
knowledge-graph like facts: subject-relation-object triples.
To test whether BERT has learned a fact, we mask the
object, thereby generating a cloze-style query, and then
evaluate predictions.

\enote{hs}{when should this seciton refer to BERT, when to PLMs?}
\enote{nk}{Let's use BERT from here on onwards}

\textbf{Symbolic reasoning.}  We create synthetic corpora to
investigate six symbolic rules (equivalence, symmetry, inversion, composition, implication, negation); see
\tabref{patterns}.
For each rule, we create a corpus that
contains facts from which the rule can be learned.  We test
BERT's ability to use the rule to infer unseen facts by
holding out some facts in a test set.  For example, for
composition, BERT should infer, after having seen that leopards
are faster than sheep and sheep are faster than snails, that
leopards are faster than snails.

Our setup is similar to link prediction in the knowledge
base domain and therefore can be seen as a natural extension
of the question: ``Language models as knowledge bases?''
\cite{petroni2019language}.  In the knowledge base domain,
prior work \cite{sun2018rotate, zhang2020learning} has shown
that models that are able to learn symbolic rules are
superior to ones that are not.

\citet{talmor2019olmpics} also investigate symbolic
reasoning in BERT using cloze-style queries.  However, in
their setup, there are two possible reasons for BERT having
answered a cloze-style query correctly: (i) the underlying
fact was correctly inferred or (ii) it was seen during
training.  In contrast, since we pretrain BERT from scratch,
we have full control over the training setup and can
distinguish cases (i) and (ii).

A unique feature of our approach compared to prior work
\cite{sinha-etal-2019-clutrr, richardson2019probing,
  DBLP:journals/corr/WestonBCM15,clark2020transformers}
is
that
we do not gather all relevant facts and present them to the
model at
inference time. This is a crucial difference -- note that
human inference similarly does not require that all relevant
facts are explicitly repeated at inference time.

We find that i) BERT is capable of learning some one-hop
rules (equivalence and implication). ii) For others,
even though high test precision suggests successful
learning, the rules were
not in fact learned correctly
(symmetry, inversion and negation).
iii) BERT
struggles with two-hop rules (composition). However,
by providing richer semantic context, even two-hop rules can
be learned.

Given that BERT can in principle learn some  reasoning
rules, the question arises whether it does so for standard
training corpora.
We find that BERT-large has only partially learned the types
of rules we investigate here.
For example, BERT has some notion of
``X shares borders with Y'' being symmetric, but it fails
to understand rules like symmetry in other cases.

\textbf{Memorization.}  During the
course of pretraining, BERT sees more data than any human
could read in a lifetime, an amount of knowledge that
surpasses its storage capacity. We simulate this 
with a scaled-down version of BERT and a 
training set that ensures that BERT cannot memorize all
facts in training. We identify two important factors
that lead to successful memorization. (i) Frequency: Other
things being equal, low-frequency facts
are not learned whereas frequent facts are. (ii) Schema
conformity:
Facts that conform with the overall schema of their entities
(e.g., ``sparrows can fly'' in a corpus with many similar
facts about birds) are easier to memorize than exceptions
(e.g., ``penguins can dive'').

We publish our code for training and data generation. \footnote{\url{https://github.com/BennoKrojer/reasoning-over-facts}}
\def\algtable{2.4cm}
\def\algtablesep{0.1cm}

\begin{figure*}[tb]
\tiny
\begin{tabular}{l@{\hspace{\algtablesep}}l@{\hspace{\algtablesep}}l@{\hspace{\algtablesep}}l@{\hspace{\algtablesep}}l@{\hspace{\algtablesep}}l}
  \begin{minipage}[t]{\algtable}
\begin{algorithmic}
\State \textbf{EQUI}
\State $C = \emptyset, D = \emptyset$
\For{$i \in 1\ldots n$}
\State $(r,s) \sim R \times R$
\State $a\sim A$
  \For{$j \in 1\ldots m$}
  \State $ e \sim E$
   \State $\mbox{addC} $=$ \mbox{Bernoulli}(0.5)$
      \If{addC}
   \State $C\! = \!C \!\cup\! \{\!(e,r,a)\!\}$
   \State $D\! = \!D \!\cup\! \{\!(e,s,a)\!\}$
   \Else
   \State $C\! = \!C \!\cup\! \{\!(e,s,a)\!\}$
      \State $D\! = \!D \!\cup\! \{\!(e,r,a)\!\}$
      \EndIf
  \EndFor
\EndFor
\end{algorithmic}
  \end{minipage}

&
  \begin{minipage}[t]{\algtable}

\begin{algorithmic}
\State \textbf{SYM}
\State $C = \emptyset, D = \emptyset$
\For{$i \in 1\ldots n$}
\State $r \sim R$
  \For{$j \in 1\ldots m$}
  \State $(e,f) \sim E \times E$
      \State $C\! = \!C \!\cup\! \{(e,r,f)\}$
      \State $D\! = \!D \!\cup\! \{(f,r,e)\}$
  \EndFor
\EndFor
\end{algorithmic}

\end{minipage}
&

  \begin{minipage}[t]{\algtable}
\begin{algorithmic}
\State \textbf{INV}
\State $C = \emptyset, D = \emptyset$
\For{$i \in 1\ldots n$}
\State $(r,s) \sim R \times R$
  \For{$j \in 1\ldots m$}
  \State $(e,f) \sim E \times E$
      \State $C\! = \!C \!\cup\! \{(e,r,f)\}$
      \State $D\! = \!D \!\cup\! \{(f,s,e)\}$
  \EndFor
\EndFor
\end{algorithmic}
  \end{minipage}
  &

  \begin{minipage}[t]{2.8cm}
\begin{algorithmic}
\State \textbf{COMP}
\State $C = \emptyset, D = \emptyset$
\For{$i \in 1\ldots n$}
\State $(r,s,t) \sim R \times R\times R$
  \For{$j \in 1\ldots m$}
  \State $(e,f,g) \sim E \times E \times E$
      \State $C\! = \!C \!\cup\! \{(e,r,f)\}$
      \State $C\! = \!C \!\cup\! \{(f,s,g)\}$
      \State $D\! = \!D \!\cup\! \{(e,t,g)\}$
  \EndFor
\EndFor
\end{algorithmic}

  \end{minipage}

&

  \begin{minipage}[t]{2.6cm}
\begin{algorithmic}
\State \textbf{IMP}
\State $C = \emptyset, D = \emptyset$
\For{$i \in 1\ldots n$}
\State $(r,s) \sim R \times R$
\For{$k \in 1\ldots l$}
\State $b \sim  A$
\State $\alpha \sim  A \times \ldots \times A$
  \For{$j \in 1\ldots m$}
  \State $ e \sim E$
      \State $C\! = \!C \!\cup\! \{(e,r,b)\}$
    \For{$a \in \alpha$}
      \State $D\! = \!D \!\cup\! \{(e,s,a)\}$
      \EndFor
    \EndFor
  \EndFor
\EndFor
\end{algorithmic}
  \end{minipage}

&

  \begin{minipage}[t]{2.7cm}
\begin{algorithmic}
\State \textbf{NEG}
\State $C = \emptyset, D = \emptyset$
\For{$i \in 1\ldots n$}
\State $r \sim R$
  \For{$j \in 1\ldots m$}
  \State $ e \sim E$
    \State $a \sim A$
    \State $b =  \mbox{antonym}(a)$
    \State $\mbox{negated} $=$ \mbox{Bernoulli}(0.5)$
      \If{negated}
           \State $C\! = \!C \cup \{\!(e,\mbox{not}\, r,a)\!\}$
      \State $D\! = \!D \!\cup\! \{(e, r,b)\}$
     \Else
      \State $C\! = \!C \!\cup\! \{(e,r,a)\}$
      \State $D\! = \!D \cup \{\!(e,\mbox{not}\, r,b)\!\}$
     \EndIf
  \EndFor
\EndFor
\end{algorithmic}
  \end{minipage}
\end{tabular}
\caption{\figlabel{genalgsym} Pseudocode for symbolic
  reasoning corpus generation.  ``$a \sim A$'' stands for:
  $a$ is randomly sampled from $A$.
(``$\alpha \sim  A \times \ldots \times A$'':
a tuple of 4 attributes is sampled.)
  The vocabulary consists of
  entities $e,f,g \in E$, relations $r,s,t \in
  R$ and attributes $a,b,c \in
  A$.
  Train/test corpora are formed from
  $C$ and $D$.
  $n=20$, $m=800$, $l=2$. See
  \secref{data_symb} for details.}
\end{figure*}

\begin{figure}[tb]
\tiny
\begin{tabular}{ll}

  \begin{minipage}[t]{3cm}
\begin{algorithmic} 
\State  \textbf{FREQ}
\State $C = \emptyset$
\State $m = 1$
\For{$i \in 1\ldots n$}
      \State $(e,f) \sim E\times E$
      \State $r \sim R$
      \For{$j \in 1\ldots m$}
      \State $C = C \cup \{(e,r,f)\}$
      \EndFor
      \If{$i\%(n/100)==0$}
      \State $ m += 1$
      \EndIf
    \EndFor
\end{algorithmic}
  \end{minipage}

&

  \begin{minipage}[t]{4.0cm}
\begin{algorithmic}
\State \textbf{SCHEMA}
\State $C = \emptyset$
\For{$i \in 1\ldots k$}
\State $ \delta \sim E \times \ldots \times E$
  \For{$r$ in $R$}
  \State $\mbox{schema} = \mbox{Bernoulli}(0.5)$
  \If{schema}
     \State $\alpha \sim A \times ... \times A$
  \For{$e \in \delta$}
  \For{$a \in \alpha$}
      \State $\mbox{add} = \mbox{Bernoulli}(0.5)$
       \If{add}
       \State $C = C \cup \{(e,r,a)\}$
       \Else
        \State $\mbox{exception} = \mbox{Bernoulli}(0.5)$
         \If{exception}
         \State $a \sim A$
          \State $C = C \cup \{(e,r,a)\}$
           \EndIf
       \EndIf
       \EndFor
    \EndFor

  \Else
     \For{$e \in \delta$}
      \State $\mbox{add} = \mbox{Bernoulli}(0.5)$
       \If{add}
   \State $a \sim A$
          \State $C = C \cup \{(e,r,a)\}$
\EndIf
   \EndFor
     \EndIf
  \EndFor
\EndFor
\end{algorithmic}
  \end{minipage}
\end{tabular}
\caption{\figlabel{pseudocode_mem} Pseudocode for
  memorization corpus generation.
  ``$a \sim A$'' stands for: $a$ is randomly sampled from $A$.
(``$ \delta \sim E \times \ldots \times E$'': a tuple of 250
  entities is sampled. ``$\alpha \sim A \times ... \times
  A$'': a tuple of 10 attributes is sampled.)
  The vocabulary consists of
  entities $e \in {E}$, relations $r \in
  {R}$ and attributes $a \in
  {A}$.
$C$ is both training set and test set. 
  $n=$ 800,000,
$k= 250$. See \secref{data_mem} for details.}
\end{figure}

\section{Data}
To test PLMs' reasoning capabilities, natural corpora like
Wikipedia are limited since it is difficult to control what
the model sees during training. Synthetic corpora provide an
effective way of investigating reasoning by giving full
control over what knowledge is seen and which rules are
employed in  generating the data.

In our investigation of PLMs as knowledge bases, it is
natural to use (subject, relation, object) triples as basic
units of knowledge; we refer to them as \emph{facts}.
The
underlying vocabulary consists of a set of entities $e,f,g,... \in {E}$, relations  $r,s,t,... \in {R}$ and attributes $a,b,c,... \in {A}$,
all represented by artificial strings such as $e_{14}$, $r_3$ or $a_{35}$.
Two types
of facts
are generated.  (i)
\textbf{Attribute facts}: relations linking entities to
attributes, e.g.,  $(e, r, a)$ = (leopard, is, fast). 
(ii) \textbf{Entity facts}: relations linking entities, e.g.,
$(e, r, f)$ = (Paris, is the capital of, France).

In the test set, we mask the objects and generate
cloze-style queries of the form ``$e$ $r$
[MASK]''. The model's task is then to predict the correct object.

\enote{nk}{is it clear that in one case the attribute is the object, in the other the enitiy?}

\subsection{Symbolic Reasoning}
\seclabel{data_symb}

\tabref{patterns} gives definitions and examples for the
six rules (EQUI, SYM, INV, COMP, IMP, NEG) we investigate.
The definitions are the basis for our corpus generation
algorithms, shown in \figref{genalgsym}.
SYM, INV, COMP generate entity facts and 
EQUI, IMP, NEG attribute facts.
We create a separate corpus for each symbolic rule.
Facts are
generated by sampling from the underlying
vocabulary.
For \secref{data_symb},
this vocabulary consists of 5000 entities, 500 relations and
1000 attributes. Half of the
relations  follow the rule, the other half is
used to generate random facts of entity or attribute type.

We can most easily think of
the corpus generation as template filling. For example,
looking at SYM in \tabref{patterns}, the template is
$(e,r, f) \iff (f,r, e)$.
We first
sample a relation $r$ from $R$ and then two entities $e$ and
$f$ from  $E$. We then add $(e,r,f)$ and $(f,r,e)$ to the
corpus -- this is one \emph{instance} of applying the SYM rule from
which symmetry can be learned. Similarly, the other rules
also generate instances.

\enote{nk}{Wo schieben wir das hin: consisting of two facts (SYM, INV, EQUI,NEG), three facts (COMP) or five facts (IMP).}

For each of the other rules, the
template filling is modified to conform with its definition
in  \tabref{patterns}.
INV corresponds directly to SYM.
COMP is a two-hop rule whereas the other five are
one-hop rules.
EQUI generates
instances from which one can learn that the relations $r$
and $s$ are equivalent. IMP generates implication instances,
e.g., $(e, r, b)$ (= (dog, is, mammal)) implies
$(e, s, a_1)$
(= (dog, has, hair)), $(e, s, a_2)$
(= (dog, has, neocortex)) etc. 
Per premise we create four implied facts.

\enote{nk}{In the table we say mammal... Maybe the distinction for attribute facts and entity facts is misleading. An attribute can also be an entity (like mammal). For attribute facts we simply sample from a different pool.}

For NEG, %
we generate pairs of facts $(e, r, a)$ (=
(jupiter, is, big)) and $(e,\mbox{not}\, r, b)$ (= (jupiter,
is not, small)). We define the antonym function
in \figref{genalgsym} (NEG)
as returning
for each attribute its antonym,
i.e., attributes are paired,
each pair consisting of a positive and a negative attribute.

Each of the six generation algorithms has
the
outer loop ``for $i$ $\in$ 1\ldots $n$'' (where $n=20$) that
samples one, two or three relations (and potentially
attributes) and generates a subcorpus for these relations;
and the
inner loop ``for $ j$ $\in$
1\ldots $m$'' (where $m=800$) that generates the subcorpus of instances for the
sampled relations.

\eat{
Filling
the respective templates follows a two step
process. 
\figref{pseudocodes} gives pseudocode. The first step fills the template
slots defining the rule. The second generates 00 instances
of that rule. Both consecutive steps are repeated 20 times,
creating 20*800 total rule instances.

These definitions serve as templates to
generate facts. The templates contain entity, relation and
attribute slots that are filled by sampling from the
underlying vocabulary. 

Accordingly, we sample 20 relations from the random relation pool and generate 800 random instances respectively. The 50 relations are sampled form the exclusively random relation pool, entities and attributes are sampled from a common pool.

Symmetry, inversion and composition are entity facts. 
First, we sample  $r_ {r} \in {R}$ to fill the relation slots defining the rule. Second, we sample  $e_ {i} \in {E}$ to fill the entity slots thereby generating multiple instances of the rule.
}

\enote{hs}{in previous versions, there was motivation for
  picking the six rule -- is that still inthe paper?}

\enote{nk}{Research on knowledge
graphs has shown that 
symmetry, inversion and composition are of central
importance for inference on knowledge graphs
\cite{guu-etal-2015-traversing, lin-etal-2015-modeling, trouillon2016complex}.}

\eat{A \textbf{symmetric rule} instance is defined by one relation $r_r \in {R}$ but two facts. We sample pairs of $e_ {i}, e_ {j} \in {E} $ and generate
$r_r(e_ {j},  e_{i})$ and $r_r(e_ {i},  e_{j})$.

An \textbf{inverse rule} instance is defined by two facts and two relations $r_r, r_l \in {R}$. We sample pairs of $e_ {i}, e_ {j} \in {E} $ and generate $r_r(e_ {j},  e_{i})$ and $r_l(e_ {i},  e_{j})$.

A  \textbf{composition rule} instance is defined by three facts and three relations $r_l$, $r_m$ and $r_n \in {R}$. We sample $e_ {i}, e_ {j},  e_ {k}\in {E} $ and generate $r_l(e_ {i},  e_{j})$, $r_m(e_ {j},  e_{k})$ and $r_n(e_ {i},  e_{k})$.}

\eat{\textbf{Equivalence} is defined by two facts, two relations and two attributes. We sample pairs of $r_k, r_l \in {R}$ and $a_k, a_l \in {A}$ to define the rule. For each pair we sample $e_ {i}, e_ {j} \in {E} $ to generate multiple instances of the rule.

For \textbf{implication}, we link a single cause fact to 5 entailed effects. Therefore, we first sample a cause and entailed $r_k \in {R}$ and a cause and 5 entailed $a_{m} \in {A}$. Second, we sample $e_ {i} \in {E} $ to generate the instances of the rule.

To model \textbf{negation} the ``not'' token is added to the vocabulary. The set of attributes ${A}$ is split in half ${A}1$ and ${A}2$. Then attributes are paired as antonyms: $a_{j}$= $antonym(a_{i})$/ $a_{i}$= $antonym(a_{j})$ for $a_{i} \in {A}1$ and $a_{j} \in {A}2$.
We sample $r_k \in {R}$, $e_ {i} \in {E} $ and  $a_ {i} \in {A}$ and generate $r_k(e_ {i},  a_{i})$ and $not$ $r_k(e_ {i},  antonym(a_{i}))$.
}

\enote{nk}{Should I add add details for negation: the antonym split?}

\textbf{Train/test split.} The data generation algorithms generate two subsets of facts $C$ and $D$, see \figref{genalgsym}. For each rule, we merge all of
$C$ with 90\% of $D$ (randomly sampled) to create the
training set. 
The rest of $D$ (i.e., the other 10\%) serves
as the test set. 

For some of the cloze queries
``$e$ $r$
[MASK]'', there are multiple correct objects that can be
substituted for MASK. Thus, we
rank predictions and compute precision at $m$, i.e.,
precision in the top $m$  where
$m$ is the number of correct objects. We average precision
at $m$ for all cloze queries.

This experimental setup allows us to test to what extent BERT learns the
six rules, i.e., to what extent the facts in the test set
are correctly inferred from their premises in the training set.

\enote{nk}{Should we justify why we made the distinction between attribute and entity facts? Perspectively we could have also had IMP, EQUI, NEG as entity facts}
\enote{nk}{In the table we say mammal... Maybe the distinction for attribute facts and entity facts is misleading. An attribute can also be an entity (like mammal). For attribute facts we simply sample from a different pool.}
\enote{nk}{Should we mention that due to sampling with negation it would be in principle be possible to generate two contradicting facts? In real data that could actually also be the case. Benno mentioned the nice example: Trump is great, Trump is terrible. We also have results based on data where we ensured that it won't happen. I am not sure which version to use! }

\subsection{Memorization}
\seclabel{data_mem}
For memorization,
the vocabulary consists of 125,000
entities, 20 relations and 2250 attributes.

\textbf{Effect of frequency on memorization.} Our first
experiment tests how the frequency of a fact influences its
successful memorization by the model.
\figref{pseudocode_mem} (left, FREQ) gives the corpus
generation algorithm.
The outer loop generates 800,000 random facts. These are
divided up in groups of 8000. A fact in the first group of
8000 is added once to the corpus, a fact from the second
group is added twice and so on. A fact from the last group
is added 100 times to the corpus.
The resulting corpus $C$ is both the training set and the
test set.

\textbf{Effect of schema conformity.}
In this experiment, we investigate 
the hypothesis  that
a fact can be memorized more easily if it is schema conformant.

\figref{pseudocode_mem} (right, SCHEMA) gives the corpus
generation algorithm. We first sample an entity group:
$\delta \sim E \times \ldots \times E$.
For each group, relations are
either related to the schema (``if schema'') or are not
(else clause).
For example, for the schema ``primate'' the relations ``eat''
(eats fruit) and ``climb'' (climbs trees) are related to the
schema, the relation ``build'' is not since some primates
build nests and treehouses, but others do not.

For non-schema relations, facts with random attributes are
added to the corpus.
In \figref{memorization}, we refer to these facts as
(facts with) \textbf{unique attributes}.
For relations related to the schema, we
sample the attributes that are part of the schema:
$\alpha \sim A \times ... \times A$ (e.g., (``paranut'',\ldots,''banana'') for ``eat'').
Facts are then generated involving these attributes and
added to the corpus.
In \figref{memorization}, we refer to these facts as
(facts with) \textbf{group attributes}.
We
also generate exceptions (e.g., ``eats tubers'') since
schemas generally have \textbf{exceptions}.

Similarly, the two lines ``$\mbox{add} =
\mbox{Bernoulli}(0.5)$'' are intended to make the data more
realistic: for a group of entities, its relations and its
attributes, the complete cross product of all facts is not
available to the human learner. For example, a corpus may
contain sentences stating that chimpanzees and baboons eat
fruit, but none that states that gorillas eat fruit.

\eat{

For the schema conform facts vs. exceptions experiments we
sample 250 entity groups. For each group we sample facts:

\begin{itemize}
 	\item  defining \textbf{common group attributes}, e.g. Can(robin, fly), Has(robin, feathers), ReproductionVia(robin, eggs), etc.
 	\item  defining \textbf{unique attributes} per entity, e.g. Has(robin, redbreast), Is(robin,  sedentary), OccurresIn(robin, Eurasia), etc.
 	\item contradicting common group attributes (\textbf{exceptions}), e.g. Can(penguin, dive). 
\end{itemize}
 
 \figref{pseudocode_mem}, SCHEMA gives the corpus generation
 algorithm.
}

For this second memorization experiment, training set and
test set are again identical (i.e., $=C$).

In a final experiment, we modify SCHEMA as
follows:
exceptions are added 10 times to the corpus (instead of
once). This tests
the interaction between schema conformity and frequency.

\eat{
frequency baseline with the scheme conformity. We use the same setup as for the scheme conformity but we repeat the exceptions 10 times whereas the facts conform with the scheme are only seen once during training.
}

\section{BERT Model}
BERT uses a deep bidirectional Transformer
\cite{NIPS2017_7181} encoder to perform masked language
modeling. During pretraining, BERT randomly masks positions
and learns to predict fillers.
We use source code provided by
\citet{Wolf2019HuggingFacesTS}.
Following \cite{DBLP:journals/corr/abs-1907-11692}, we perform
dynamic masking and no next sequence prediction.

For symbolic rules, we start with BERT-base
and tune hyperparameters. We vary the number of layers to avoid that rule learning fails due to over-parametrization, see appendix for details.
We report precision based on optimal configuration.

In the memorization experiment, our goal is to investigate
the effect of frequency on memorization.
Due to a limited
compute infrastructure, we scale down BERT to a single hidden layer with 3 attention
heads, a hidden size of 192 and an intermediate size of 768.

\begin{table}
\small
\centering
\begin{tabular}{l|r|r}
rule &train &test  \\\hline\hline
EQUI & 99.95&98.28 \\
SYM & 99.97 &98.40 \\
INV & 99.99&87.21\\
IMP & 100.00 & 80.53 \\
NEG & 99.98 &20.54 \\
COMP  & 99.98&0.01\\\hline
ANTI & 100.00 &14.85 
\end{tabular}
\caption{\tablabel{symresults} Precision in \% of completing
  facts for symbolic rules. Training corpora
  generated as specified in
  \figref{genalgsym}.
  See \secref{results}
  for  detailed discussion.}
\end{table}

\begin{figure}
\centering
\begin{tabular}{cc}
 \includegraphics[width=0.45\linewidth]{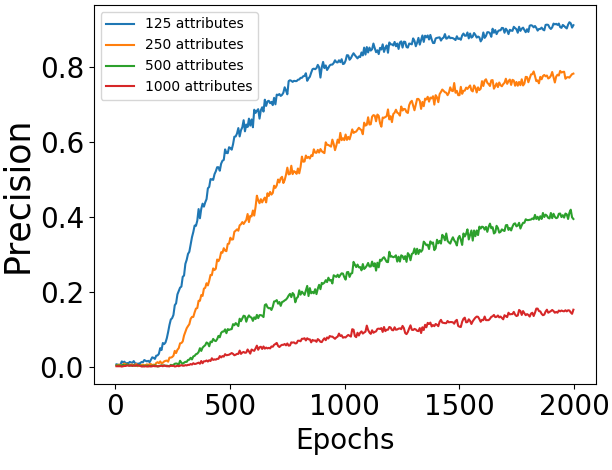}&\includegraphics[width=0.45\linewidth]{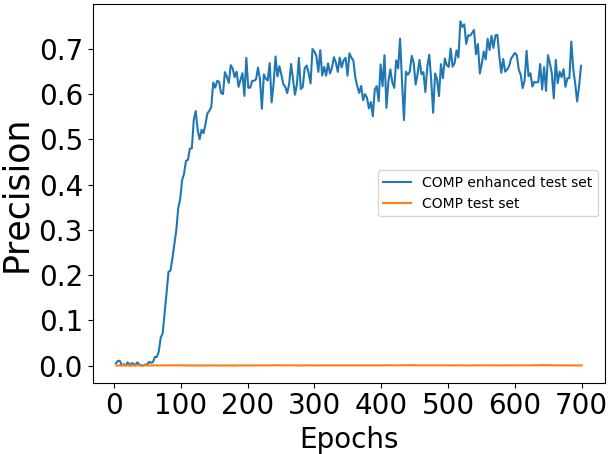} \\
   (A) NEG & (B) COMP\\
\end{tabular}
\caption{\figlabel{learningcuves} 
Learning curves for symbolic reasoning. (A) shows precision
for NEG with a varying number of attributes. A reduction to
125 attributes enables BERT to successfully apply antonym
negation to the test set. (B) shows test set precision for
COMP following the standard setup (orange) and an enhanced
version (blue). Only in enhanced, i.e., with the
introduction of  additional facts adding more semantic
information, is COMP  generalized.}
\end{figure}

\section{Results, Analysis and Discussion}

\subsection{Symbolic Reasoning}
\seclabel{results}
\tabref{symresults} gives results for the symbolic reasoning
experiments. BERT has high test set precision for
{EQUI}, {SYM},
{INV} and {IMP}.
As we see in \tabref{patterns},
these rules share that they
are ``one-hop'': The inference can be
straightforwardly made from a single premise to a
conclusion, e.g., 
``(barack married michelle)'' implies ``(michelle married
barack)''. The crucial difference to prior work is
that the premise is not available at inference
time. ``(michelle married
barack)'' is correctly inferred by the model based on its
memory of having seen the fact ``(barack married michelle)''
in the training set and based on the successful acquisition
of the symmetry rule. \tabref{symresults}  seems to suggest that
BERT is able to learn one-hop rules and it
  can successively apply these rules in a natural setting in
  which the premise is not directly available.

In the rest of this section, we investigate these results
further for SYM, INV, NEG and COMP.

\eat{
For {NEG} model parameter tuning was necessary.  \figref{learningcuves} C) shows how a reduction to 6 layers improves results. Still, NEG is not fully learned. Double the number of rules, the vocabulary and rule instances thereby generating a four times larger training corpus. Still, negation is not fully learned. 
By tweaking design-parameters such as total number of
attributes, we see an indication of generalization.}

\subsubsection{Analysis of SYM and INV}
\tabref{symresults} seems to indicate that BERT can learn that a
relation $r$ is symmetric (SYM) and that $s$ and $t$ are inverses
(INV) -- the evidence is that it generates facts
based on the successfully acquired symmetry and inversion properties of
the relations 
$r$, $s$ and $t$. We now show that while BERT
acquires SYM and INV partially, it also severely
overgenerates. Our analysis points to the complexity of
evaluating rule
learning in PLMs and opens interesting avenues for future
work.

Our first observation is that in the SYM experiment, BERT
understands \emph{all relations} to be symmetric. Recall
that of the total of 500 relations, 250 are symmetric and
250 are used to generate random facts. If we take a fact
with a random relation $r$, say $(e,r,f)$, and prompt BERT
with ``$(f,r,[MASK])$'', then $e$ is predicted in close to
100\% of cases. So BERT has simply learned that any relation
is symmetric as opposed to distinguishing between symmetric
and non-symmetric relations.

This analysis brings to light that our setup is unfair to
BERT: it never sees evidence for non-symmetry.
To address this, we define a new experiment, which we call
ANTI because it includes an additional set of ``anti'' relations that are sampled from
$R^{*}$ with
$R^{*}$$\cap$ $R$ =$\emptyset$ and $|R| = |R^{*}|$. %
ANTI facts
take the following form:
$(e, r, f)$, $(f, r, g)$ with $e \neq
g$. Using this ANTI template we follow the standard data generation procedure.
The corpus is now composed of symmetric, anti-symmetric and random facts.
ANTI training data indicate to BERT that $r \in R^{*}$ is not
symmetric since many instances of $r$ facts are seen, with
specific entities ($f$ in the example) occurring in both
slots, but there is never a symmetric example.

\tabref{symresults} (ANTI) shows that BERT memorizes ANTI
facts seen during training but on test,
BERT only recognizes 14.85\% of ANTI facts as non-symmetric.
So it still generalizes from the 250
symmetric relations to most  other relations (85.15\%), even those
without any ``symmetric'' evidence in training. So it is  easy for
BERT to learn the concept of symmetry, but it is  hard
to teach it to distinguish between symmetric and
non-symmetric relations. 

Similar considerations apply to INV. BERT successfully
predicts correct facts once it has learned that $s$ and
$t$ are inverses -- but it overgeneralizes by also
predicting many incorrect facts; e.g., for $(e,s,f)$ in
train, it may predict $(f,t,e)$ (correct), but also
$(e,t,f)$ and $(f,s,e)$ (incorrect).

In another INV experiment, we 
add, for each pair of $(f,r,e)$ and $(e,s,f)$
two facts that give evidence of  non-symmetry:
$(f,r,g)$ and $(e,s,h)$ with $e \neq
g$ and $h \neq f$.
We find that test set precision for INV (i.e., inferring
$(e,s,f)$ in test from $(f,r,e)$ in train) drops to 17\% in
this scenario. As for SYM, this indicates how complex the
evaluation of rule learning is.

In summary, we have found that SYM and INV are learned in
the sense that BERT generates correct facts for symmetric
and inverse relations. But it severely overgenerates.
Our analysis points to a problem of neural language models
that has not received sufficient attention: they can
easily learn that the order of
arguments is not important (as is the case for SYM relations), but it is hard for them to
learn that this is the case \emph{only for a subset of
  relations}. Future work will have to delineate the exact
scope of this finding -- e.g., it may not hold for much
larger training sets with millions of occurrences of each
relation. 
Note, however, that human learning is likely to have a
bias against symmetry in relations since the vast majority
of verbs\footnote{For example, almost all of the verb
  classes in
\cite{lev93} are asymmetric.}
  in English (and presumably relations in the world)  is asymmetric.
So unless we have explicit
evidence for symmetry, we are likely to assume a relation is
non-symmetric. Our results suggest that neural language
models do not have this bias -- which would be problematic
when using them for learning from natural language text.

\subsubsection{Analysis of NEG}
NEG was the only rule for which
parameter tuning improved performance. A reduction to
four layers obtained optimal results.

In \tabref{symresults} we report a test set precision of
20.54\%.  Why is negation more challenging than implication?
Implication allows the model to generalize over several
entities all following the same rule (e.g., every animal
that is a mammal has a neocortex). This does not hold for
negation (e.g., a leopard is fast but a snail is not
fast). BERT must learn antonym negation from a large number
of possible combinations. By reducing the number of possible
combinations (decreasing the number of attributes from 1000
to 500, 250 and 125) BERT's test set precision increases,
see \figref{learningcuves} (A). With 125 attributes a
precision of 91\% is reached. A reduction of attributes
makes antonym negation very similar to implication.

We investigate BERT's behavior concerning negation further
by adding an additional attribute set $A^{*}$, with $A^{*}$
$\cap$ $A$ =$\emptyset$ and $|A| = |A^{*}|$ to the vocabulary.  $A^{*}$ does not
follow an antonym schema. We sample $a \in A^{*}$, $e \in
E$, $r \in R$ to add additional random facts of the type
($e$, $r$, $a$) or ($e$, \mbox{not} $r$, $a$) to NEG's training
set. After training we test on the additional random facts
seen during training by inserting or removing the negation
marker. We see that BERT is prone to predict both ($e$, $r$, $b$)
and ($e$, \mbox{not} $r$, $b$) for b $\in$ $A^{*}$ (for 38\%). Antonym negation was still learned.

We conclude that {antonym negation can be learned via
  co-occurrences but a general concept of negation is not
  understood.}

This is in agreement with prior work \cite{BERTNOT,
  Kassner2020NegatedAM} showing that BERT trained on natural
language corpora is as likely  to generate a true
statement like ``birds can fly'' as a  factually false
negated statement like ``birds cannot fly''.

 \begin{table*}
 \small
 \centering
\begin{tabular}{l|c|rrr|l}
relation & rule & \multicolumn{3}{c|}{completions} &
examples\\
&&cons. & correct & inc.&\\
\hline\hline
\emph{shares borders with} & SYM &  152&152 & 2 &
\begin{tabular}[b]{@{}l@{}}\textcolor{blue}{(ecuador,peru)}\\\textcolor{red}{(togo,ghana),
    (ghana,nigeria)}\end{tabular}\\
\hline
\emph{is the opposite of} & SYM & 179&170 & 71 &
\begin{tabular}[b]{@{}l@{}}\textcolor{blue}{(demand,supply)}\\
  \textcolor{red}{(injustice,justice), (justice,truth)}\end{tabular}\\
\hline
\begin{tabular}[b]{@{}l@{}}\emph{is the capital of
    (C-of)}\\\emph{'s capital is (s-C-is)}\end{tabular}& INV &
59&59& 1 &
\begin{tabular}[b]{@{}l@{}}\textcolor{blue}{(indonesia,s-C-is,jakarta)}\\\textcolor{red}{(canada,s-C-is,ottawa),
    (ottawa,C-of,ontario)}\end{tabular}\\
\hline
\hline
\begin{tabular}[b]{@{}l@{}}\emph{is smaller/larger than}\\(countries)\end{tabular}& INV &  54&23& 99&
\begin{tabular}[b]{@{}l@{}}\textcolor{blue}{(russia,larger,canada),
    (canada,smaller,russia)}\\\textcolor{red}{(brazil,smaller,russia),
    (russia,smaller,brazil)}\end{tabular}\\
\hline
\begin{tabular}[b]{@{}l@{}}\emph{is smaller/larger than}\\(planets)\end{tabular}& INV &  9&9& 36&
\begin{tabular}[b]{@{}l@{}}\textcolor{blue}{(jupiter,larger,mercury), 
(mercury,smaller,jupiter)}\\\textcolor{red}{(sun,bigger,earth), (earth,bigger,sun)}\end{tabular}
\end{tabular}
\caption{\tablabel{nl}
Can PLMs (BERT and RoBERTa) learn SYM and INV from natural
language corpora? For ``smaller/larger'', we follow
\citet{talmor2019olmpics} and test which of the two words is
selected as a filler in a pattern like ``Jupiter is [MASK]
than Mercury''. For the other three relations, we test
whether the correct object is predicted (as in the rest of
the paper). We give the number of (i) consistent 
(``cons.''), (ii) correct and consistent (``correct'')  and (iii) inconsistent (``inc.'') predictions.
Blue: consistent examples. Red: inconsistent examples. (We
make the simplifying assumption that ``justice'' can only
have one opposite.)}
\end{table*}

\subsubsection{Analysis of COMP}
Why does BERT not learn COMP?
COMP differs from the other rules in that it involves
two-hop reasoning. Recall that a novelty of our experimental
setup is that premises are not presented at inference time
-- two-hop reasoning requires that two different facts have
to be ``remembered'' to make the inference, which
intuitively is harder than a one-hop inference.
\figref{learningcuves} (B)
shows that the problem is not 
undertraining (orange line).

Similar to the memorization experiment, we investigate
whether stronger semantic structure in form of a schema can
make COMP learnable. We refer to this new experiment as
\textbf{COMP enhanced}.
Data generation is defined as
follows: Entities are divided into groups of 10. Relations
are now defined between groups in the sense that the members
of a group are ``equivalent''. More formally, we sample
entity groups (groups of 10)
$E_1$,
$E_2$, $E_3$ and relations $r$, $s$, $t$. For all $e_1 \in
E_1, e_2 \in E_2, e_3 \in E_3$,
we add 
$(e_1,r,e_2)$ and $(e_2,s,e_3)$ to
$C$ and $(e_1,t,e_3)$ to $D$. In addition, we introduce a relation
``samegroup'' and add, for all $e_m,e_n \in E_i$,
$(e_m,\mbox{samegroup},e_n)$ to $C$ -- this makes it easy to
learn group membership.
As before, the training set is
the merger of $C$ and 90\% of $D$ and the test set is the
rest of $D$.

Similar semantic structures occur in real data. The simplest
case is a transitive example: ($r$) planes (group 1) are faster than
cars (group 2), ($s$) cars (group 2) are faster than bikes
(group 3), ($t$) planes (group 1) are faster than bikes
(group 3).

\figref{learningcuves} (B) shows that BERT can learn COMP
moderately well from this schema-enhanced
corpus (blue curve): precision is clearly above 50\% and
peaks at 76\%.

The takeaway from this experiment is that two-hop rules pose
a challenge to BERT, but that they are learnable if
entities and relations are embedded in a
rich
semantic structure. Prior work
\cite{brown2020language}
has identified the absence of
``domain models'' (e.g., a domain model for common sense
physics) as one
shortcoming of PLMs. To the extent that PLMs lack such
domain knowledge (which we simulate here with a schema),
they may not be able to learn COMP.

\subsection{Natural Language Corpora}
In this section, we investigate to what extent
the PLMs BERT and RoBERTa have learned SYM and INV from natural
language corpora.
See \tabref{nl}.
For ``smaller/larger'' (INV), we follow
\citet{talmor2019olmpics} and test which of the two words is
selected as the more likely filler in a pattern like ``Jupiter is [MASK]
than Mercury''. For the other three relations
(``shares borders with'' (SYM), ``is the opposite of'' (SYM), ``is the
capital of'' / ``'s capital is'' (INV)), 
we test
whether the correct object is predicted
in the pattern ``$e$ $r$ [MASK]''
(as in the rest of
the paper). We give the number of (i) consistent 
(``cons.''), (ii) correct and consistent (``correct'') and (iii) inconsistent (``inc.'') predictions. (A
prediction is consistent and incorrect if it is consistent
with the rule, but factually incorrect.)

In more detail, we take a set of entities
(countries like ``Indonesia'', cities like
``Jakarta'') or adjectives like ``low'' that are 
appropriate for
the relation and test which of the entities / adjectives is
predicted. For each of the five relations, we run 
both BERT-large-cased and RoBERTa-large and 
report the more consistent result. 

Consistency and accuracy are high for
``shares borders with'' and ``capital''. However, this is
most likely due to the fact that many of these facts occur
verbatim in the training corpora of the two models.
For example, Google shows 54,800 hits for
``jakarta is the capital of indonesia''
and 1,290 hits for ``indonesia's capital is jakarta''
(both as a phrase). It is not possible to determine
which factor is decisive here: successful rule-based
inference or memorization. The ultimate futility of this
analysis is precisely the reason that we chose to work with
synthetic data.

Consistency for ``is the opposite of'' is much lower than
for the first two relations, but still decent.
To investigate this relation further, we also tested the
relation ``is the same as''. It turns out that many of the
``opposite'' objects are also predicted for ``is the same
as'', e.g., ``high is the same as \emph{low}'' and ``low is the same as
\emph{high}'' where the predicted word is in italics.
This indicates that the models have not 
really learned that ``is the opposite of'' is symmetric, but
rather know that antonyms are closely associated and often
occur together in phrases like 
``X is the
opposite of Y'',
``X and Y'', ``X noun, Y noun'' (e.g., ``good cop, bad
cop'') etc. Apparently, this is then incorrectly generalized
to ``is the same as''.

Consistency and accuracy are worse for ``smaller/larger''.
``smaller/larger'' sentences of the sort considered here are
probably rarer in genres like Wikipedia than ``shares
borders with'' and ``is the capital of''. A Wikipedia
article about a country will always say what its capital is
and which countries it borders, but it will not enumerate
the countries that are smaller or larger.

In summary, {although we have shown that pretrained
  language models have some ability to  learn symbolic
  rules, there remains considerable doubt that they can
  do so based on natural corpora.}

\begin{figure*}
\centering
\begin{tabular}{ccc}
\includegraphics[width=0.3\linewidth]{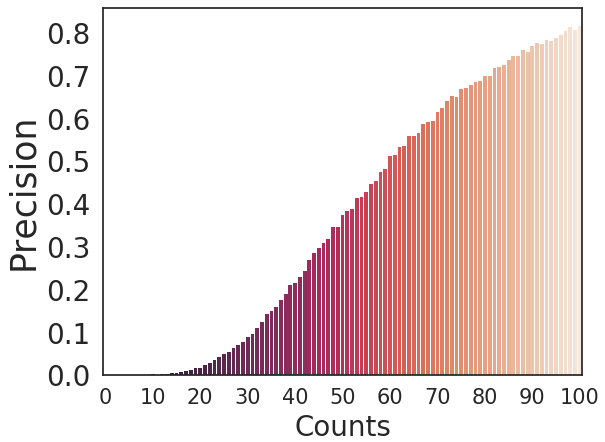} &  \includegraphics[width=0.3\linewidth]{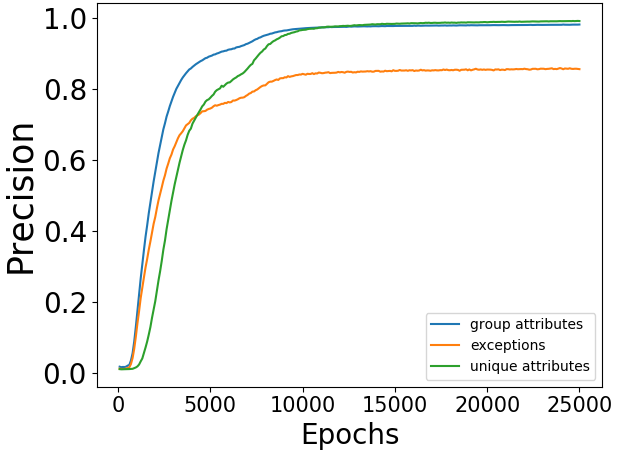} &\includegraphics[width=0.3\linewidth]{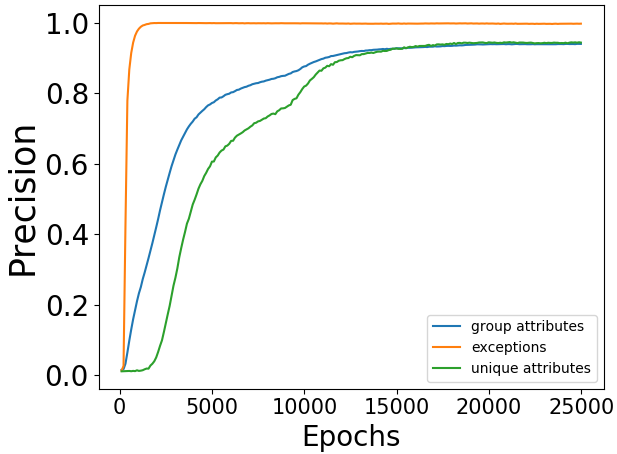}\\
(A) frequency & (B) schema conformity,  & (C) schema conformity,  \\
 & exceptions rare  & exceptions frequent  \\
\end{tabular}
\caption{\figlabel{memorization} Memorization experiments.
  We investigate the effect of frequency and schema
  conformity on memorization. (A) Frequent facts are
  memorized well (0.8 for frequency 100), rare facts are not
  ($\approx$ 0.0 for frequencies $<15$).
  (B) BERT memorizes schema conformant facts perfectly
  (``group attributes''). Accuracy for rare exceptions is clearly
  lower (80\%).
  (C)
  Exceptions are perfectly learned if 10
  copies of each exception are added to the corpus -- instead
  of 1 in (B). In this case,
limited capacity affects memorization of schema-conformant
facts (``group attributes'' drops to $\approx$ 0.9).}
\end{figure*}

\subsection{Memorization}
Experimental results for the memorization experiments are shown in \figref{memorization}.

(A) shows that frequent facts are
  memorized well (0.8 for frequency 100) and that rare facts are not
  ($\approx$ 0.0 for frequencies $<15$).
  
  (B) shows that BERT memorizes schema conformant facts perfectly
  (``group attributes'').
  Accuracy for exceptions is clearly
  lower than those of schema conformant facts: about 80\%.
The frequency of
each fact in the training corpus in this experiment is 1. Overall, the total amount of exceptions
is much lower than the total amount of schema conformant facts.
  \eat{
In the beginning of training BERT
focuses on attributes common for individual groups as well
as exceptions going against those group attributes. In the
course of training BERT picks up on group membership and
unique facts distinguishing entities. Mean precision for the
exceptions plateau when BERT's storage capacities are maxed
out. All other facts conform with the overall schema reach a
mean precision of 100 \%.
  }
  
(C)  shows that
  exceptions are perfectly learned if 10
  copies of each exception are added to the corpus -- instead
  of 1 in (B). In this case,
limited capacity affects memorization of schema-conformant
facts: accuracy  drops to $\approx$ 0.9.

In summary, we find that both {frequency and schema
  conformity facilitate memorization}.
Schema conformant facts and
exceptions compete for memory if memory capacity is limited
-- depending on frequency one or the other is preferentially
learned by BERT.

\eat{
i'm not sure i agree and/or i'm not sure this is new information

(e.g., ``bert has incentive to store exceptinos if seen
frequently enough'' is redundant with the simple fact that
higher frequency makes memorizaiton more likely)

We see that fact frequency is the dominant factor
influencing if a fact is stored or forgotten. Considering
real world factual knowledge involving exceptions this seems
desirable as many world concepts and categories go along
with exceptions. Exceptions are not factually false e.g. a
penguin can dive but not fly. BERT has the incentive to
store exceptions if seen frequent enough during
training. For rare facts not deducible via a semantic schema
this incentive is missing.

}

\eat{

In C) we show results for the combination of the frequency baseline with the semantic scheme. We again report learning curves for the respective types of facts. Right after the first training epoch, BERT's prediction accuracies for the frequent exceptions jumps high and plateaus at 1.0 early on in training. The other learning curves pick up slowly. Again the group attributes are stored earlier than the other facts conform with the scheme. Towards the end of training the preference of frequent facts even tough they contradict the overall scheme remains unchanged. This time facts conform with the overall scheme are forgotten and exceptions are stored.

}
\section{Limitations}
Our experimental design makes many simplifying assumptions:
i) Variation in generated data is more limited than in
naturally occurring data. ii)
Semantics
are deliberately restricted to one rule only per generated corpus.
iii)
We do not investigate effects of  model and corpus size.

i) In natural corpora relations can have more than two
arguments, entities can have several tokens, natural data
are noisier than synthetic data etc. Also, we study each
rule in isolation.

\enote{nk}{should we add here that design parameters do influence results: proportion relations following rules, amount of entities etc.}

ii)
While our simplified corpora make learning easier in some
respects, they may make it harder in others. Each corpus is
focused on providing training material for one symbolic
rule, but it does not contain any other ``semantic'' signal
that may be helpful in learning symbolic reasoning:
distributional signals,
entity groupings, hierarchies, rich context etc.
The experimental
results of ``COMP enhanced'' indicate that indeed such
signals are beneficial to symbolic rule learning. The
interplay of such additional sources of information for
learning with symbolic rules is an interesting
question for follow up work.

iii) Results are based on BERT-base and scaled-down versions
of BERT-base only, just as training corpora are orders of
magnitude smaller than natural training corpora.  We varied
model and corpus sizes within the limits of our compute infrastructure,
but did not systematically study  their effect on our
findings.  \enote{nk}{should we add here that indeed for NEG
  we saw an effect of model size on findings but for all
  other rules we did not?}

Our work is an initial exploration of the question whether
symbolic rules can be learned in principle, but
we view it mainly as a starting point for future work.

\section{Related Work}
\citet{radford2019language}
and \citet{petroni2019language}
show
in a zero-shot question answering setting
that PLMs
have factual knowledge.
Our main question is: under what conditions do PLMs learn
factual knowledge and do they do so through memorization or
rule-based inference?

\citet{sun2018rotate} and \citet{zhang2020learning} show in the
knowledge graph domain that models that have
the ability to capture symbolic rules like SYM, INV and COMP
outperform ones that do not. We investigate this question
for PLMs that are trained on language corpora.

\eat{
In contrast, we pose the question whether BERT is generally able to learn symbolic rules during pretraining.
}

\citet{talmor2019olmpics}
 test PLMs' symbolic reasoning
capabilities probing pretrained and finetuned models with
cloze-style queries. 
Their setup makes it impossible to distinguish
whether a fact was inferred or memorized during pretraining.
Our synthetic corpora allow us to make this distinction.

\citet{clark2020transformers} test finetuned BERT's reasoning capabilities, but they always
make premise and conclusion locally available to the model, during training and inference. 
This is arguably not the way much of human inference works;
e.g.,
the fact $F$ that X borders Y allows us to infer that Y
borders X even if we were exposed to $F$ a long time ago.

\citet{richardson2019probing} introduce
synthetic corpora testing logic and monotonicity
reasoning. They show that BERT performs poorly on these new
datasets, but can be quickly finetuned to good
performance. The difference to our work again is
that they make the
premise available to the model at inference time.

\eat{maybe not necssary?
  
We
analyze BERT's pretraining objective, ask for multi-hop rule
learning and do not state the rules explicitly.
}

For complex reasoning QA benchmarks
\cite{yang-etal-2018-hotpotqa, sinha-etal-2019-clutrr}, PLMs
are finetuned to the downstream tasks. Their performance is
difficult to analyze: it is not clear whether any reasoning
capability is learned by the PLM or by the task specific
component.

Another line of work \cite{gururangan-etal-2018-annotation,
  kaushik-lipton-2018-much, Dua2019DROPAR, McCoy2019RightFT}
shows that much of PLMs' performance on reasoning tasks is
due to statistical artifacts in datasets and does not
exhibit true reasoning and generalization
capabilities. With the help of synthetic corpora,
we can cleanly investigate PLMs' reasoning capabilities.

\citet{Hupkes2020CompositionalityDH} study 
the ability of neural models to capture compositionality.
They do not investigate our six rules, nor do they
consider the effects of fact frequency and schema conformity.
Our work confirms their finding that
transformers have
the ability to capture both rules and
exceptions. 

A large body of research in psychology and cognitive science
has investigated how some of our rules are processed in
humans, e.g., \citet{Sloman1996-SLOTEC-2} for
implication. There is also a lively debate in cognitive
science as to  how important rule-based reasoning is for  human
cognition 
\cite{Politzer2007-POLRWC}.

\citet{yanaka-etal-2020-neural, goodwin-etal-2020-probing} are concurrent studies of systematicity in PLMs. The first shows that monotonicity inference
is feasible for syntactic structures close to the ones observed during training. The latter shows that PLMs can exhibit high over-all performance on natural language inference despite being non-systematic.

\citet{Roberts2020HowMK} show that the amount of knowledge
captured by PLMs increases with model size. Our
memorization experiments investigate the factors that
determine successful acquisition of knowledge.

\citet{Guu2020REALMRL} modify the PLM objective to
incentivize knowledge acquisition. They do not consider symbolic rule learning nor do they analyze what factors influence successful memorization.

Based on perceptrons and convolutional neural networks, 
\citet{10.5555/3305381.3305406, DBLP:journals/corr/ZhangBHRV16} study the relation of generalizing from real structured data vs. memorizing random noise in the image domain, similar to our study of schema-conformant facts and outliers. They do not study transformer based models trained on natural language.

\section{Conclusion}
We studied
BERT's  
ability to capture knowledge
from its training corpus
by
investigating its reasoning and memorization
capabilities. We identified factors influencing what
makes successful memorization possible
and what is
learnable beyond knowledge explicitly seen during training.
We saw that, to some extent,  BERT is able to infer facts not
explicitly seen during training via symbolic rules.

Overall, effective knowledge acquisition must combine both
parts of this paper: memorization and symbolic reasoning. A
PLM is not able to store an unlimited amount of
knowledge. Through acquiring reasoning
capabilities, knowledge gaps can be filled based on
memorized facts. A schema-conformant
fact (``pigeons can fly'')  need not be memorized if
there are a few facts that indicate that birds fly and then
the ability of flight can be filled in for the other birds. The
schema conformity experiments suggest that this is
happening. It is easier to capture knowledge that conforms with a
schema instead of memorizing facts one by one.

There are several directions for future work.
First, we made many simplifying assumptions that should be
relaxed in future work.
Second, how can we improve PLMs' ability to learn symbolic
rules? We see two avenues here, either additional inductive
biases could be imposed on PLMs' architectures or training
corpora could be modified to promote learning of symbolic rules.

\section{Acknowledgements}
We thank Peter Clark for helpful discussions and our reviewers for constructive feedback.

This work was funded by the German Federal Ministry of Education
and Research (BMBF, Grant No. 01IS18036A)
and by Deutsche Forschungsgemeinschaft (DFG, Grant ReMLAV:
Relational Machine Learning for Argument Validation). The authors of this work take full responsibility for its content.

\bibliographystyle{acl_natbib}
\bibliography{anthology,symbolic_reasoning}
\newpage
\cleardoublepage

\appendix
\section{Hyperparameters}
\subsection{Model hyperparameters}
For all reported results we trained with a batch-size of 1024 and a learning rate of 6e-5.

Our experiments for symbolic rules started with the BERT-base model with 12 layers, 12 attention heads, hidden size of 768 and intermediate size of 3072. For rules with a low test precision (NEG and COMP) we then conducted a restricted grid search (restricted due to limited compute infrastructure): We tried all possible numbers of layers from 1 to 12 and then only considered the best result. For NEG the best performance came from 4 layers, whereas COMP did not show improvements for any number of layers. For NEG with 3 layers (which had a very similar performance to 4 layers) we exemplarily tested whether changing the attention heads, hidden size or intermediate size improves precision. For this we trained with the following 4 settings:

\begin{itemize}
\item attention heads = 6, hidden size = 768, intermediate size = 3072
\item attention heads = 12, hidden size = 384, intermediate size = 1536
\item attention heads = 12, hidden size = 192, intermediate size = 768
\item attention heads = 12, hidden size = 96, intermediate size = 192
\end{itemize}

However this did not further improve precision.
\subsection{Data hyperparameters}
In previous iterations of our experiments, we had used different settings for generating our data. For instance, we had varied the number of rules in our corpora: 50 or 100 instead of the presented 20 rules. Even the sampling process itself can be tweaked to allow for less overlaps between rules and between instances of one rule. However, we observed the same trends and similar numbers across these different settings.
\newpage
\section{Symbolic rules}
In the following sections, we present illustrating corpora for \textbf{INV}, \textbf{IMP} and \textbf{COMP enhanced}. Each line is one datapoint. We also include the control group at the end of each corpus that does not follow any rule.
In the case of composition enhanced, "\{...\}" indicates the sampled group which is not part of the actual dataset.

We illustrate our training corpora using real world entities and relations. Note that the actual corpora used for training are composed of an entirely synthetic vocabulary. For simplicity we show grouped composition with enhancement with groups of 4, instead of 10 as it is in the real data.
\subsection{INV}
\noindent Paris CapitalOf France

\noindent France HasCapital Paris

\noindent ...

\noindent Egypt HasCapital Cairo (counterpart in test-set)

\noindent ...

\noindent Apple Developed iOS

\noindent iOS DevelopedBy Apple

\noindent ...

\noindent Germany RandomRelation China

\noindent Cairo RandomRelation Norway

\subsection{IMP}
\{(Flu), (Cough, RunningNose, Headache, Fever)\}

\noindent Kevin HasDisease Flu

\noindent Kevin HasSymptom Cough

\noindent  Kevin HasSymptom RunningNose

\noindent Kevin HasSymptom Headache

\noindent Kevin HasSymptom Fever

\noindent ...

\noindent Mariam HasDisease Flu

\noindent ...

\noindent Peter RandomRelation Pain

\noindent Sarah RandomRelation Tooth

\newpage
\subsection{Comp enhanced}

\{e8, e2, e4, e5\}

\noindent e8 ConnectedTo e2

\noindent e8 ConnectedTo e4

\noindent e8 ConnectedTo e5

\noindent e2 ConnectedTo e8

\noindent ...

\noindent \{e15, e13, e12, e19\}

\noindent e15 ConnectedTo e13

\noindent e15 ConnectedTo e12

\noindent ...

\noindent \{e25, e24, e29, e20\}

\noindent e25 ConnectedTo e24

\noindent e25 ConnectedTo e29

\noindent ...

\noindent e8 r1 e15

\noindent e8 r1 e13

\noindent e8 r1 e12

\noindent e8 r1 e19

\noindent e2 r1 e15

\noindent e2 r1 e13

\noindent ...

\noindent e5 r1 e19

\noindent ...

\noindent e15 r2 e25

\noindent e15 r2 e24

\noindent e15 r2 e29

\noindent e15 r2 e20

\noindent ...

\noindent e19 r2 e20

\noindent ...

\noindent e8 r3 e25

\noindent e8 r3 e24

\noindent e8 r3 e29

\noindent e8 r3 e20

\noindent ...

\noindent ...

\noindent e133 r61 e23

\noindent e56 r61 e29

\noindent ...

\end{document}